\documentclass[letterpaper, 10 pt, conference]{ieeeconf}  

\IEEEoverridecommandlockouts                              

\usepackage{cite}
\usepackage{amsmath,amssymb,amsfonts}
\usepackage{algorithmic}
\usepackage{graphicx}
\usepackage{textcomp}
\usepackage{xcolor}
\usepackage{float}
\usepackage{lineno,hyperref}
\begin{document}

\title{Asynchronous Event-Inertial Odometry using a Unified Gaussian Process Regression Framework}

\author{
	Xudong Li, 
	Zhixiang Wang, Zihao Liu,
	Yizhai Zhang, Fan Zhang, Xiuming Yao
	and Panfeng Huang
	\thanks{This work was supported by the National Natural Science Foundation of China under Grant 62022067.  (\emph{Corresponding author: Yizhai Zhang. e-mail:zhangyizhai@nwpu.edu.cn.})}
	\thanks{Xudong Li, Yizhai Zhang, Fan Zhang and Panfeng Huang are with Shaanxi Province Innovation Team of Intelligent Robotic Technology, School of Astronautics, Northwestern Polytechnical University, Xi’an, China.}
	\thanks{Zhixiang Wang and Zihao Liu are with Shaanxi Province Innovation Team of  Intelligent Robotic Technology, School of Automation, Northwestern Polytechnical University, Xi'an, China.}
	\thanks{Xiuming Yao is with the School of Electronic and Information Engineering, Beijing Jiaotong University, Beijing, China.}
}

\maketitle

\begin{abstract}
Recent works have combined monocular event camera and inertial measurement unit to estimate the $SE(3)$ trajectory. However, the asynchronicity of event cameras brings a great challenge to conventional fusion algorithms. In this paper, we present an asynchronous event-inertial odometry under a unified Gaussian Process (GP) regression framework to naturally fuse asynchronous data associations and inertial measurements. A GP latent variable model is leveraged to build data-driven motion prior and acquire the analytical  integration capacity. Then, asynchronous event-based feature associations and integral pseudo measurements are tightly coupled using the same GP framework. Subsequently, this fusion estimation problem is solved by underlying factor graph in a sliding-window manner. With consideration of sparsity, those historical states are marginalized orderly. A twin system is also designed for comparison, where the traditional inertial preintegration scheme is embedded in the GP-based framework to replace the GP latent variable model. Evaluations on public event-inertial datasets demonstrate the validity of both systems. Comparison experiments show competitive precision compared to the state-of-the-art synchronous scheme. 
\end{abstract}

\section{Introduction}

The event camera is a bio-inspired visual sensor that has an underlying asynchronous trigger mechanism where 
the illumination intensity variety of the scenario is independently recorded in a per-pixel manner. Benefiting from this special mechanism, event cameras gain prominent performance improvements to conventional cameras, such as low power consumption, low latency, high dynamic range, high temporal resolution, etc\cite{gallego2020event}. 
In the field of robotic state estimation, the event camera belongs to the exteroceptive sensor, and the Inertial Measurement Unit (IMU) is regarded as a proprioceptive sensor. 
As the proprioceptive sensor and the exteroceptive sensor  naturally have strong complementarity, recent works have applied the combination of them to further enhance the performance of robotic state estimation in aggressive motion.

Typically, the frame-based feature tracking scheme, IMU preintegration\cite{forster2016manifold} and discrete-time fusion methods should be effective to estimate robotic state. The frame-based feature tracking accumulates events within a fixed time window (or a certain quantity) to generate intensity frames called event frames. 
As the event frame has abandoned the temporal diversity of events within the same frame, motion blur will occur, and additional deblurring operations need to be applied by introducing motion compensation \cite{vidal2018ultimate}. Then, traditional feature detector and tracking methods are used to search for inter-frame data associations. Consequently, the high temporal resolution of event cameras will be degenerated. Other frame-like accumulation schemes also exist the same problem \cite{guan2022monocular}. After that, the robotic motion will be estimated at sparse discrete temporal points by discrete-time fusion methods, such as Extented Kalman Filter (EKF) and Bundle Adjustment. Similarly, within this discrete framework, traditional IMU preintegration is applied to construct a relative motion constraint between two keyframes. Since a mass of inter-frame temporal measurements is abandoned, the fine-grained motion information will be lost as well. 
Therefore, it's imperative to introduce novel methods for fusing asynchronous and high-temporal-resolution observation of event-inertial odometry. 

\begin{figure}[!t]
	\centering
	\includegraphics[width=3.4in]{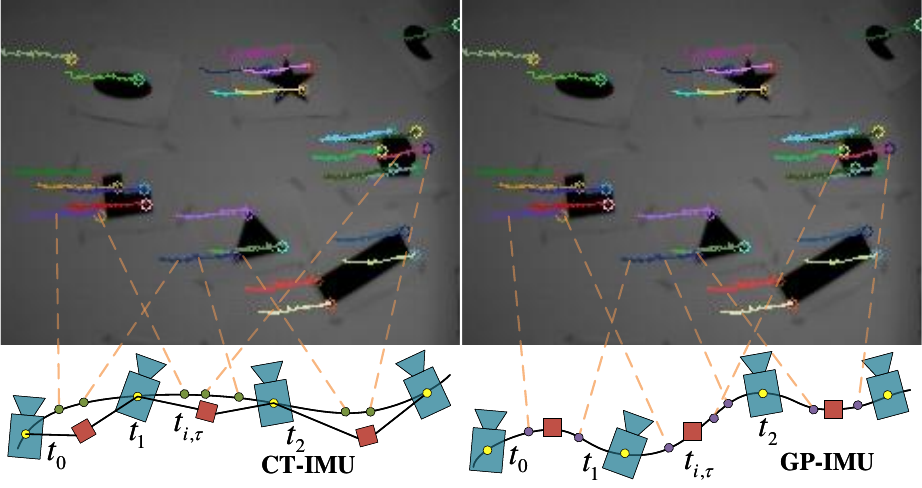}
	\caption{Illustration of the two asynchronous measurements fusing ways. Left is the Continuous-time Trajectory with standard IMU preintegration (CT-IMU) method, which queries state (green dot) on CT and fuses IMU measurements (red block) separately. Right is the Gaussian Process IMU preintegration (GP-IMU) method, relying on IMU measurements for state inference. Gray image visualizes the feature tracking.}
	\label{featuretraj}
\end{figure}

Recently, continuous-time estimation approaches are designed to support the state inference from asynchronous measurements like Fig. \ref{featuretraj}. 
GP-based continuous-time methods have recently been studied to alleviate the aforementioned issues of discrete-time scheme for event-inertial odometry. 
Nevertheless, the performance comparison between IMU-induced GP and IMU preintegration has not been studied under the same visual front-end and GP framework.
In addition, previous works used to solve the GP optimization problem using a full-smoother back-end, which is very time-consuming. 

In this paper, we focus on asynchronous measurements fusing way problem and propose an asynchronous event-inertial odometry system using a unified GP-based fusion framework. 
On the visual front-end, an event-driven feature detecting and tracking method called EKLT\cite{gehrig2020eklt} is adopted to construct fully asynchronous data associations and maintain the high temporal resolution of event cameras. Meanwhile, a latent variable model is introduced to the GP regression framework to analytically integrate inertial measurements and implicitly induce a data-driven GP. On the back-end, IMU-induced GP and asynchronous data associations are concurrently built into the factor graph as normal measurement factors. 
A marginalization strategy proposed in our previous
 work \cite{wang2024efficient} is further leveraged to maintain the underlying sliding-window factor graph by removing historical states under consistency guarantees. A twin system, based on IMU preintegration, is further accomplished for comparison under the same visual front-end and GP regression framework.

\section{Related Works}
From the foregoing, event-inertial odometry can be partitioned  into two categories, i.e., frame-like discrete-time schemes and event-driven continuous-time methods. Inheriting algorithms from conventional frame-based cameras, the frame-like methods perform data associations on the event accumulation such as event frame\cite{2017Real,vidal2018ultimate} or Time Surface (TS) \cite{guan2022monocular}. Similarly, they fuse data associations and inertial measurements in a discrete-time optimization or filtering manner. Rebecq et al.\cite{2017Real} detected the features on motion-compensated edge event image and tracked through Lucas Kanade (LK), and then fused the IMU-preintegrated measurements. Guan et al.\cite{guan2022monocular} used two different TS to perform robust feature tracking and tightly fused the event-corner features with IMU data under a standard keyframe-based framework. Zhu et al. \cite{zihao2017event} combined the feature tracks from batch event packets with the output of IMU by an EKF back-end. Overall, the methods mentioned above treat events as batching way, ignoring the high temporal resolution characteristics of event cameras.  

To unleash the potential of event cameras, recent researchers have focused on directly processing event streams in an event-by-event manner\cite{clady2015asynchronous,alzugaray2020haste, gehrig2020eklt, hu2022ecdt,dai2022tightly}. Subsequently, a continuous-time back-end\cite{mueggler2018continuous,liu2022asynchronous} is adopted to naturally model the asynchronous data associations. In front-end, HASTE\cite{alzugaray2020haste} proposed a multi-hypothesis to update features from every new event. Dai et al.\cite{dai2022tightly} applied an exponential decay kernel to associate asynchronous corners extracted by Arc*\cite{alzugaray2018asynchronous}. Other works also used line feature representation \cite{mueggler2018continuous,le2020idol,chamorro2023event} to track asynchronously. To cope with the asynchronous tracking, Mueggler et al.\cite{2015Continuous} firstly used B-spline trajectory representation and optimized the control points. Later, this work is extended with IMU\cite{mueggler2018continuous} to improve the accuracy and recover the scale. 

Gaussian Process (GP) is another continuous-time representation way early used in inferring the motion trajectory with the scanning lidar and unsynchronized sensors \cite{dong2018sparse, wong2020data, wang2023event}. Liu et al.\cite{liu2022asynchronous} firstly applied the GP continuous-time trajectory estimation in monocular event-based visual odometry (VO) system with a white-noise-on-acceleration (WNOA)\cite{anderson2015full} sparse prior. \cite{wang2024efficient} implemented a monocular event-based visual system based on GP continuous-time trajectory and investigated dynamic marginalization strategy. Nevertheless, the sparse prior needs the trajectory to satisfy certain assumptions such as constant velocity, acceleration, or recent data-driven model\cite{wong2020data}. With IMU measurements, Gentil et al.  \cite{le2020gaussian} formulated the acceleration as GP and upsample the gyroscope measurements which allowed to interpolate pose relying on discrete state. This asynchronous fusing paradigm was applied in a line-based VIO system to \cite{le2020idol,dai2022tightly}. However, these systems have high computational costs for their growing full-batch optimization or partial sliding window optimization\cite{le2020idol}. Extended from \cite{le2020gaussian}, latent state was used to formulate a more unified GP representation way in \cite{le2023continuous}.  Based on\cite{wang2024efficient}, we leverage the GP representation for IMU\cite{le2023continuous} to mitigate the motion prior constraint. Besides, we also apply the traditional preintegration to our origin system for comparison.
\section{Methodology}
\subsection{Overview}\label{subsec1}
\begin{figure}[!t]
	\vspace{8pt}
	\centering
	\includegraphics[width=3.5in]{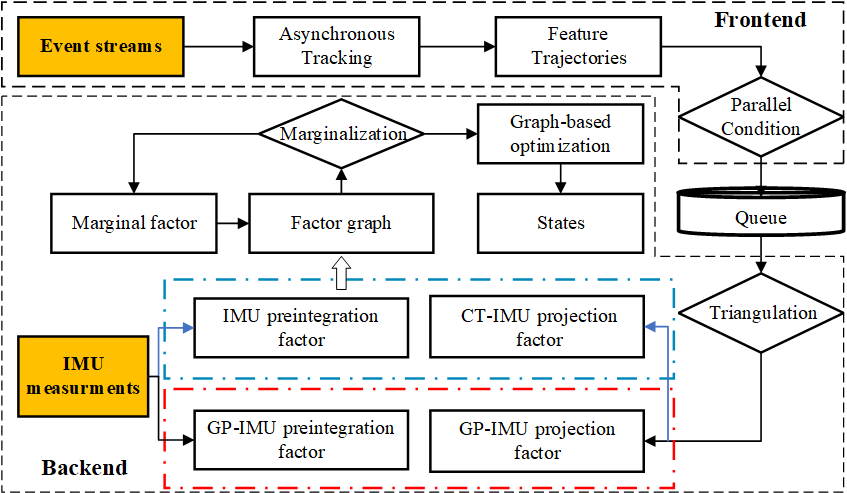}
	\caption{System pipeline. The whole system consists of two essential modules. The front-end process the raw events to associated feature trajectories in parallel. The back-end here uses two different GP-based ways(highlight by two blocks) fusing asynchronous measurements.}
	\label{system_pipeline}
\end{figure}

The framework of the proposed asynchronous event-inertial odometry system is illustrated in Fig \ref{system_pipeline}. It primarily consists of two parallel threads: 1) asynchronous feature tracking front-end, and 2) GP-based sliding-window back-end. Event streams serve as inputs to the front-end for asynchronous detection and tracking to get feature trajectories. Afterwards, the feature manager assigns a global key to each feature trajectory and selectively pushes those that satisfy the parallax threshold to the back-end. For asynchronous feature trajectories, the back-end firstly query the pose at the measurement time, then attempt to triangulate the new feature and add those triangulated measurements to the graph as projection factors. Meanwhile, the inertial measurements are used to predict the initial value of state and to construct IMU preintegration. Eventually, all these associated visual observations, IMU preintegration and marginalized prior are modeled as factors added to a graph and optimized under the sliding-window manner (\ref{subsec6}).

Now, we declare frame definitions and notations throughout the paper. The world frame is denoted by $w$, and the direction of gravity is aligned with the z-axis of this frame. The extrinsic transformation from the camera frame $c$ to the body (IMU) frame $b$ is noted as $\boldsymbol{T}_{bc}$ and is kept constant in our estimation. $\boldsymbol{R}_{bc}$  and $\boldsymbol{p}_{b}^{cb}$ denote the rotation matrix and position vector in $\boldsymbol{T}_{bc}$. To simplify the symbol, if not specified, $(\cdot)_{b}^{cb}$ represents in begin frame $b$ is omitted its subscript as the $(\cdot)^{cb}$. We denote $\hat{(\cdot)}$ as the measurement of a certain quantity. 

\subsection{Sparse GP Priors}\label{subsec2}
Sparse GP prior maintains a sparse structure to efficiently interpolate and optimize. The main idea of this method is to model the local state from linear time-invariant stochastic differential equations as sparse GP. We follow the previous work\cite{wang2024efficient} using the WNOA motion prior for $SE(3)$ as follows:
\begin{equation}
	\label{GPprior without linear}
	\begin{aligned}
		\dot{\boldsymbol{T}}_{wb}(t) &= \boldsymbol{T}_{wb}(t)\boldsymbol{\varpi}_b^{bw}(t)^\land,
		\\
		\dot{\boldsymbol{\varpi}}_b^{bw}(t) &= \boldsymbol{w}(t),\quad \boldsymbol{w}(t) \sim \boldsymbol{GP}(0,\boldsymbol{Q}_c \delta(t-t')),
	\end{aligned}
\end{equation}
where $\boldsymbol{\varpi}_b^{bw}(t) = [\boldsymbol{\mathbf{v}}_b^{bw}(t), \boldsymbol{\mathbf{w}}_b^{bw}(t)]^T \in \mathbb{R}^6$ is the body-frame velocity, $(\bullet)^\land$ maps a element in $\mathbb{R}^6$ to $\mathfrak{se}(3)$,  $\boldsymbol{w}(t)$ is the generalized acceleration vector modeled as a zero-mean, white-noise GP, $\boldsymbol{Q}_c$ is a
usual power spectral density matrix, and $\delta (t-t')$ is Dirac’s delta function. To consistent with IMU-preintegration state in \ref{subsec3}, the motion state we estimated is $\boldsymbol{x}_m(t) \triangleq \{\boldsymbol{T}_{wb}(t),\boldsymbol{\varpi}^{bw}(t)\}$, which can be directly transformed to $\boldsymbol{\varpi}_b^{bw}(t)$ by left product the $\boldsymbol{T}_{wb}(t)$.

Using the transformation between global state and local state, the prior residual between the adjacent discrete motion states pair $\boldsymbol{x}_{m_k}, \boldsymbol{x}_{m_{k+1}}$ can be written as:
\begin{equation}
	\label{PriorRes}
	\boldsymbol{r}_{\mathcal{P}}(\boldsymbol{x}_{m_k},\boldsymbol{x}_{m_{k+1}}) = \left[\begin{array}{cc}
		(t_{k+1}-t_k)\boldsymbol{\varpi}_k-log(\boldsymbol{T}_k^{-1}\boldsymbol{T}_{k+1})^\vee\\
		\boldsymbol{\varpi}_k-\boldsymbol{J}_r(log(\boldsymbol{T}_k^{-1}\boldsymbol{T}_{k+1})^\vee)^{-1}\boldsymbol{\varpi}_{k+1}
	\end{array}
	\right],
	\end{equation}
	where $log(\bullet)$ is the map from $SE(3)$ to $\mathfrak{se}(3)$ and $\boldsymbol{\varpi}_k$ is the $t_{k}$ body-frame velocity in \eqref{GPprior without linear}.

	\subsection{IMU Preintegration}\label{subsec3}
	IMU preintegration was proposed to avoid recompute integration when the linearization point changes by partially integrating the gyroscope and accelerometer measurements to a relative motion increment form $\hat{\boldsymbol{z}}_{b_k b_{k+1}} = [\Delta\hat{\boldsymbol{R}}_{b_k b_{k+1}}, \Delta\hat{\boldsymbol{v}}^{b_{k+1} {b_k}}, \Delta\hat{\boldsymbol{p}}^{b_{k+1} {b_k}}]^T$: 
	\begin{equation}
		\begin{aligned}
		\label{IMU preinte}
		   \Delta\hat{\boldsymbol{R}}_{b_k b_{k+1}} &= \prod_{t_k}^{t_{k+1}}exp((\hat{\boldsymbol{\omega}}(t)-\boldsymbol{b}_\omega(t))^\land dt),
		   \\
		   \Delta\hat{\boldsymbol{v}}^{b_{k+1} {b_k}} &= \int_{t\in[t_k, {t_{k+1}}]}\Delta\boldsymbol{R}_{b_k b}(\hat{\boldsymbol{a}}(t)-\boldsymbol{b}_a(t))dt,
			\\
			\Delta\hat{\boldsymbol{p}}^{b_{k+1} {b_k}} &= \iint_{t\in[t_k, {t_{k+1}}]}\Delta\boldsymbol{R}_{b_k b}(\hat{\boldsymbol{a}}(t)-\boldsymbol{b}_a(t))dt.
		\end{aligned} 
	\end{equation}
	Define the IMU state as $\boldsymbol{x}_k\triangleq\{\boldsymbol{T}_{wb_k}, \boldsymbol{v}^{b_kw}, \boldsymbol{b}_{a_k}, \boldsymbol{b}_{\omega_k}\}$, the IMU residual errors $\boldsymbol{r}_{\mathcal{I}}(\boldsymbol{x}_k,\boldsymbol{x}_{k+1}, \hat{\boldsymbol{z}}_{b_k b_{k+1}} ) = [ \boldsymbol{r}_{\Delta \boldsymbol{R}}, \boldsymbol{r}_{	\Delta\boldsymbol{v}}, \boldsymbol{r}_{	\Delta\boldsymbol{p}},
	\boldsymbol{r}_{	\Delta\boldsymbol{b}} ]^T$, each term is:
	\begin{equation}
		\label{IMU preinte res}
		\begin{aligned}
			\boldsymbol{r}_{\Delta \boldsymbol{R}} &= log(\Delta\hat{\boldsymbol{R}}_{b_k b_{k+1}}^T \boldsymbol{R}_{b_k{b_{k+1}}}^T)^\vee,
			\\
			\boldsymbol{r}_{\Delta\boldsymbol{v}} &= \boldsymbol{R}_{wb_k}^T(\boldsymbol{v}^{b_{k+1}w}-\boldsymbol{v}^{b_{k}w}-\boldsymbol{g}\Delta t_{k,k+1} )
			- \Delta\hat{\boldsymbol{v}}^{b_{k+1} {b_k}},
			\\
			\boldsymbol{r}_{\Delta\boldsymbol{p}} &= \boldsymbol{R}_{wb_k}^T(\boldsymbol{p}^{b_{k+1}w} - \boldsymbol{p}^{b_{k}w} - \boldsymbol{v}^{b_{k}w}\Delta t_{k,k+1} 
			\\
			&- \frac{1}{2}\boldsymbol{g}\Delta t_{k,k+1}^2 ) - \Delta\hat{\boldsymbol{p}}^{b_{k+1} {b_k}},
			\\
			\boldsymbol{r}_{\Delta\boldsymbol{b}} &= \boldsymbol{b}_{k+1}-\boldsymbol{b}_{k}
		\end{aligned}
	\end{equation}
	where $\boldsymbol{R}_{b_k{b_{k+1}}}^T$ is the relative rotation
matrix between two IMU states. Here, we omit bias correction in \eqref{IMU preinte res} to simplify the formula. The bias is kept  constant between two discrete times to avoid repeating the integration. A complete form of IMU preintegration is provided in\cite{forster2016manifold}.
	\subsection{GP Regression Preintegration}\label{subsec5}

	The continuous latent state preintegration models the relative acceleration and rotation vector's velocity as independent GP. With the linear operator and inference ability of GP, it can query the given time preintegrate state from the latent state measurements. To perform linear inference, we model the rotation vector's derivation and relative acceleration as GP:
	\begin{equation}
		\label{UGPM GP}
		\begin{aligned}
			\dot{\boldsymbol{r}}_{b_k b}(t) \sim \boldsymbol{GP}(0,\boldsymbol{k}_r(t,t')),
			\\
			\boldsymbol{a}^{b b_k}(t) \sim \boldsymbol{GP}(0,\boldsymbol{k}_a(t,t')),
		\end{aligned}
	\end{equation}	  
	where $\boldsymbol{a}^{b b_k}(t)$ is the acceleration of time $t$ in $b_k$ frame, $\dot{\boldsymbol{r}}_{b_k b}(t)$ is the rotation vector's derivation and $\boldsymbol{k}_r(t,t'),\boldsymbol{k}_a(t,t')$ is the respective kernel function. To get the noisy observations of the modeled GP, we introduce the latent state $\hat{\boldsymbol{\rho}}$ and $\hat{\boldsymbol{\alpha}}$ at IMU measurements time $t_i$:
	\begin{equation}
		\label{UGPM meas}
		\begin{aligned}
			\hat{\boldsymbol{\rho}}_i = \dot{\boldsymbol{r}}_{b_k b_i} + \boldsymbol{\eta}_r \ with \  \boldsymbol{\eta}_r \sim \mathcal{N}(0,\boldsymbol{\sigma}_r^2),
			\\
			\hat{\boldsymbol{\alpha}}_i = \boldsymbol{a}^{b_i b_k} + \boldsymbol{\eta}_a \ with\  \boldsymbol{\eta}_a \sim \mathcal{N}(0,\boldsymbol{\sigma}_a^2),
		\end{aligned}
	\end{equation}
	where $\boldsymbol{\eta}$ is zero mean Gaussian distribution. Then, an optimization problem is constructed to estimate the latent states using IMU measurements, with details provided in\cite{le2023continuous}.
	
	With the inference and linear operator of GP, the preintegrated velocity $\Delta \boldsymbol{v}^{b_{\tau} b_k}$, position $\Delta \boldsymbol{p}^{b_{\tau} b_k}$and rotation vector increment $\Delta \boldsymbol{r}_{b_k b_{\tau}}$ can be written as:
	\begin{equation}
		\label{Query UGPM}
		\begin{aligned}
			\Delta \boldsymbol{r}_{b_k b_{\tau}} &= \mathcal{L}_r^t k_r(t_{\tau},\boldsymbol{t})[\boldsymbol{K}_r(\boldsymbol{t},\boldsymbol{t})+\sigma_r^2\boldsymbol{I}]^{-1}\hat{\boldsymbol{\rho}},
			\\
			\Delta \boldsymbol{v}^{b_{\tau} b_k} &= \mathcal{L}_v^t k_a(t_{\tau},\boldsymbol{t})[\boldsymbol{K}_a(\boldsymbol{t},\boldsymbol{t})+\sigma_a^2\boldsymbol{I}]^{-1}\hat{\boldsymbol{\alpha}},
			\\
			\Delta \boldsymbol{p}^{b_{\tau} b_k} &= \mathcal{L}_p^t k_a(t_{\tau},\boldsymbol{t})[\boldsymbol{K}_a(\boldsymbol{t},\boldsymbol{t})+\sigma_a^2\boldsymbol{I}]^{-1}\hat{\boldsymbol{\alpha}},
		\end{aligned}
	\end{equation}
	where $\mathcal{L}_p^t$, $\mathcal{L}_v^t$ and $\mathcal{L}_r^t$ are the integral linear operator as described in \cite{le2023continuous}, $t_{\tau}$ is the query time. The given time's relative rotation increment is calculated by exponential map:

	\begin{equation}
		\label{Tlinear}
		\Delta\hat{\boldsymbol{R}}_{b_k b_{\tau}} =  \exp(\Delta \boldsymbol{r}_{b_k b_{\tau}}).
	\end{equation}

	\subsection{Sliding Windows Graph based Optimization}\label{subsec6}
	\begin{figure}[thpb]
		\vspace{8pt}
		\centering
		\includegraphics[width=3.4in]{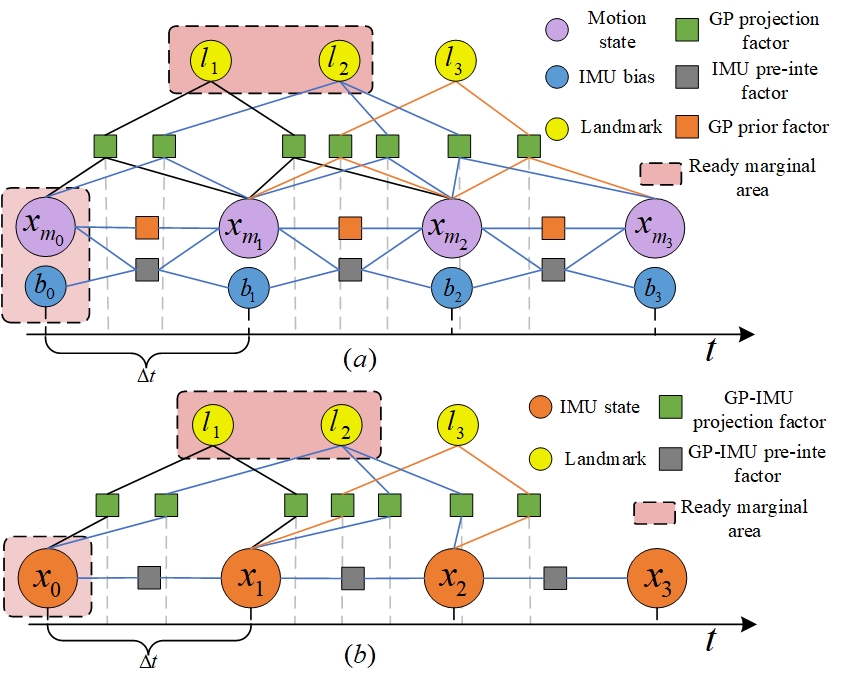}
		\caption{Factor graph of two fusing ways. (a) is the factor graph of sparse GP prior + IMU preintegration (CT-IMU) method, (b) is the GP regression preintegration method (GP-IMU). Note the IMU state in \ref{subsec3} contains bias term. The ready marginal area is the marginalization order of our graph.}
		\label{factor_graph}
	\end{figure}
	\subsubsection{Interpolation projection factor}
	Interpolation on sparse GP prior is done through local linear state\cite{dong2018sparse}. In contrast, the interpolation on GP of IMU data is performed by inference on latent states \eqref{Query UGPM}. From the above interpolation operation, we obtain the position and rotation at measurement time $t_{\tau}$ as $\boldsymbol{T}_{wb_\tau}$. Using the landmark representation, it is now easy to write the visual residual error as $\boldsymbol{r}_{\mathcal{V}} \in \mathbb{R}^3$, where:
	\begin{equation}
		\label{Visual residual UGPM}
		\boldsymbol{r}_{\mathcal{V}}(\boldsymbol{T}_{wb_\tau}, \boldsymbol{l}_{i}, \hat{z}_i) = \hat{z}_i-\frac{1}{d_i}\boldsymbol{K}(\boldsymbol{T}_{wb_\tau}\boldsymbol{T}_{b_\tau c_\tau})^T\boldsymbol{l}_{i},
	\end{equation}
	$\boldsymbol{l}_{i}$ is the landmark position in world frame, $\hat{z}_i$ is the visual measurement on pixel plane and $\boldsymbol{K}$ is the intrinsic matrix of camera.  
	\subsubsection{Sparse GP Prior + IMU preintegration (CT-IMU)}\label{method1} 	
	 The collection of all states in the sliding-window at current time $t_c$ defined as $\boldsymbol{\chi}_c$:
	\begin{equation}
		\label{state}
		\boldsymbol{\chi}_c = [\boldsymbol{x}_{m_0},\boldsymbol{x}_{m_1},...\boldsymbol{x}_{m_{n-1}}, \boldsymbol{b}_0, \boldsymbol{b}_1,...,\boldsymbol{b}_{n-1},\boldsymbol{l}_0,\boldsymbol{l}_1,...\boldsymbol{l}_{m-1}]^T,
	\end{equation}
	where $\boldsymbol{x}_{m_0}$ is the motion state defined in \ref{subsec2}, $\boldsymbol{b}_i$ is the respectively bias state.
	
	Our optimizer minimizes the Mahalanobis norm of GP prior residuals \eqref{PriorRes}, IMU preintegration residuals \eqref{IMU preinte res}, GP interpolated projection residuals \eqref{Visual residual UGPM} and marginal residuals to obtain a maximum posterior estimation as:
	\begin{equation}
		\begin{split}
			\label{CTMAP}
			\mathop{\min}_{\chi_c}\{\sum_{k\in K}&(\Vert \boldsymbol{r}_{\mathcal{P}} \Vert_{\boldsymbol{Q}_{k+1}}^2 + \Vert \boldsymbol{r}_{\mathcal{I}} \Vert_{\boldsymbol{\Sigma}_{b_{k+1}}^{b_k}}^2)\\
			&+\sum_{z_{l_i}\in L}e(\Vert \boldsymbol{r}_{\mathcal{V}} \Vert_{\boldsymbol{\Sigma}_\tau}^2)+\Vert \boldsymbol{r}_{\mathcal{M}} \Vert\ \},
		\end{split}
	\end{equation}
	$\boldsymbol{Q}_{k+1}$, $\boldsymbol{\Sigma}_{b_{k+1}}^{b_k}$ and $\boldsymbol{\Sigma}_\tau $ represents the covariance matrix. $\boldsymbol{r}_{\mathcal{P}}$, $\boldsymbol{r}_{\mathcal{I}}$ and $\boldsymbol{r}_{\mathcal{V}}$ are defined in previous sections. $e(\bullet)$ is the Huber norm. $\Vert \boldsymbol{r}_{\mathcal{M}} \Vert$ is the marginal prior residual calculated from schur complement.
	\subsubsection{GP Regression Preintegration (GP-IMU)}\label{method2} 
	GP-IMU reduce the prior on trajectory and interpolate the needed state related to previous IMU state and GP preintegration measurement. The different of two estimation problem in factor graph is shown in Fig. \ref{factor_graph}. The MAP problem is:
	\begin{equation}
		\begin{split}
			\label{GPMAP}
			\mathop{\min}_{\boldsymbol{\chi}_{c}'}\{\sum_{k\in K}&\Vert \boldsymbol{r}_{\mathcal{I}} \Vert_{\boldsymbol{\Sigma}_{b_{k+1}}^{b_k}}^2 +\sum_{z_{l_i}\in L}e(\Vert \boldsymbol{r}_{\mathcal{V}} \Vert_{\boldsymbol{\Sigma}_\tau}^2)+\Vert \boldsymbol{r}_{\mathcal{M}} \Vert\ \},
		\end{split}
	\end{equation}
	where states $\boldsymbol{\chi}_{c}'$ contains the landmark 
	 $\boldsymbol{l}$ and IMU state $\boldsymbol{x}$ defined in \ref{subsec3}. 
	\subsubsection{Marginalization for sliding window}\label{Marginalization}
	In frame-based odometry system, sliding window is widely used to bounded the scale of the optimization problem by performing the marginalization for keyframe and discarding non-keyframe. However, the associated visual measurements used in our system can not be seen as frame. So we apply a dynamic marginalization, by firstly marginalizing the latest state and related landmarks like in Fig. \ref{factor_graph} the ready marginal area to maintain the Hessian matrix sparse shape, the detail we refer the reader to see\cite{wang2024efficient}. After marginalization, the induced marginal distribution serves as a prior factor applied to the new factor graph as in \eqref{GPMAP}.
	\section{Evaluation}
	\subsection{Implementation Details}\label{subsec4.1}  
	
	The whole event-inertial odometry system is implemented in C++. The event streams and inertial measurements are subscribed from standard ROS topics. 
	For the front-end, the \emph{EKLT}  needs intensity frames solely for feature detection and to build the initial template patch. Therefore, the feature trajectory tracked by \emph{EKLT} is still asynchronous and retains the high temporal resolution.
	A factor-graph solver called \emph{GTSAM}\footnote{https://github.com/borglab/gtsam} is integrated into the system to implement the GP-based sliding-window back-end and to solve the optimization problem. 
	The back-end is an extension of  our previous implementation \cite{wang2024efficient}, which achieves higher solving efficiency and robustness than the full-smoothing method \cite{liu2022asynchronous}.

	In our experiments, the parallax condition is set to $8$ pixel, and the minimum of feature trajectories is set to $4$ for front-end. The number of latent states for GP-IMU is set to $400$, which is lower than the frequency of IMU measurements. $Q_c$ in \eqref{GPprior without linear} is set to an identity diagonal matrix as $0.05$. For the back-end, the minimum window size is set to 40 and the time interval between two states is set to 0.05 seconds. In our test, we find that it's a reasonable choice, though these parameters can be tuned according to the specific situation. 
	
	\subsection{Experimental Evaluation}\label{subsec4.2}
	To demonstrate the performance of our proposed system, we evaluate our two methods on several sequences from the DAVIS\cite{mueggler2017event} and MVSEC\cite{zhu2018multivehicle} datasets. To compare with discrete methods, we conduct evaluations of  \cite{vidal2018ultimate} and the raw trajectory from \cite{guan2022monocular}. The DAVIS dataset is recorded using DAVIS240C(240*180) in multiple scenes with aggressive motion. For the MVSEC dataset, we only use data from the left event camera(DAVIS 346B, 346×260). In comparison, we adapt the event and IMU(E+I) configuration for \cite{vidal2018ultimate}.

	\begin{figure}[!t]
		\vspace{8pt}
		\centering
		\includegraphics[width=3.0in]{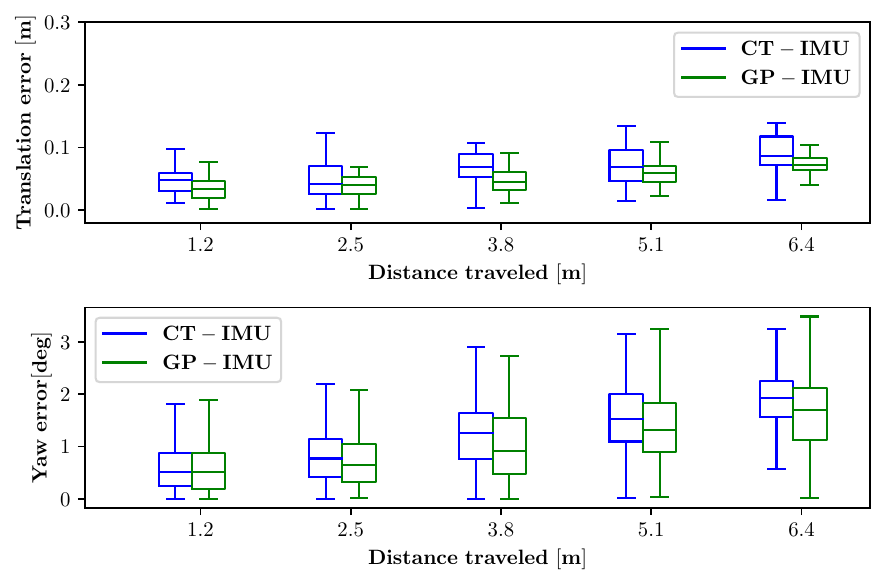}
		\caption{The relative errors of the translation (top) and yaw angle (bottom) in the first 40 s using the different algorithms upon dynamic\_6dof sequence.}
		\label{dynamic_6dof_error}
	\end{figure}
	\begin{figure}[!t]
		\centering
		\includegraphics[width=3.5in]{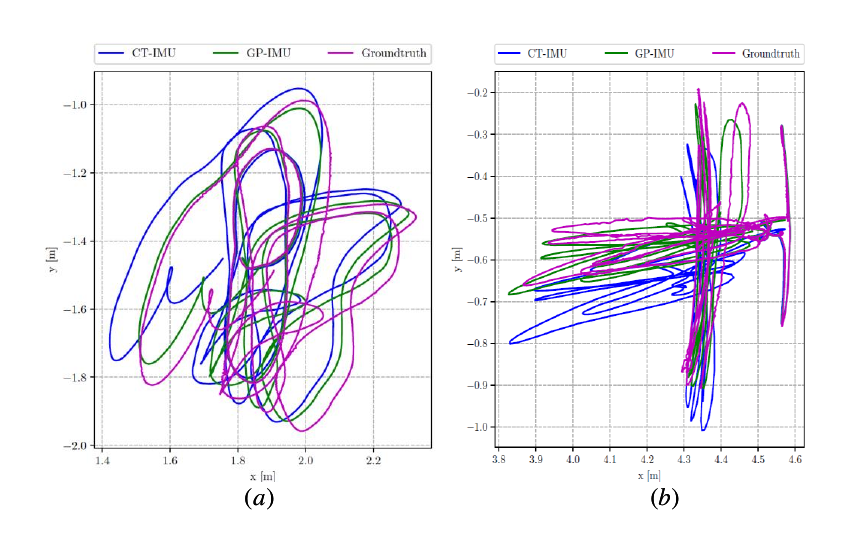}
		\caption{(a) Estimated trajectory of dynamic\_6dof  aligned with ground truth. (b) Estimated trajectory of poster\_translation.}
		\label{dynamic_6dof_traj}
	\end{figure}
	To quantitatively analyze our results, we align the ground-truth and estimated trajectories using the first 5 seconds of trajectory. The root mean square relative trajectory errors (RMS RTE) of the aligned trajectory are used to evaluate accuracy. 
	\begin{table}[htbp]
		
		\caption{The RMS RTE of each method on various sequences }
		\begin{center}
			\vspace{-2.0em}
			\begin{tabular}{c|c|c|c|c}
				\hline\hline
				Sequences& \textbf{\textit{CT-IMU}} & \textbf{\textit{GP-IMU}} & Ref.\cite{guan2022monocular} & Ref.\cite{vidal2018ultimate}(E+I)\\
				\hline
				dynamic\_translation& \textbf{0.030} & 0.060 & 0.056 & 0.037 \\
				
				dynamic\_6dof& 0.076 & 0.056 & 0.073 & \textbf{0.040} \\
				
				poster\_translation& 0.087  & \textbf{0.082} & 0.242 &  0.087 \\
				
				poster\_6dof& 0.156  & \textbf{0.084} & 0.210 & 0.197 \\
				
				boxes\_6dof& 0.347  & 0.151 & \textbf{0.073} &  0.078 \\
				
				shapes\_6dof& \textbf{0.108}  & 0.244 & -\--\-- & 0.163  \\
				
				indoor\_flying1& 2.167  & \textbf{1.506} &   -\--\-- & 1.968 \\
				
				indoor\_flying2& 2.278  & \textbf{2.149} &   -\--\-- & failed \\
				\hline
			\end{tabular}
			\vspace{-2.0em}
			\label{RMSE_table}
		\end{center}
	\end{table}

	\subsubsection{Accuracy}\label{accuracy}
	The visualization of estimated results on two sequences is depicted in Fig.\ref{dynamic_6dof_traj}. Intuitively, the estimated results are very close to the ground truth trajectory. The trajectory of CT-IMU is smoother than GP-IMU which can be explained by its GP prior constraining the velocity change rate. The translation error and yaw error corresponding to the trajectory segment of the total trajectory length are shown in Fig.\ref{dynamic_6dof_error}. Table \ref{RMSE_table} presents the RMS RTE on different sequences for our two methods and \cite{vidal2018ultimate,guan2022monocular}. Generally, the precision of our two methods is competitive with the discrete optimization method and outperforms in some sequences. However, since our methods do not include outlier exclusion or motion-compensation in the front-end like \cite{vidal2018ultimate,guan2022monocular}, the system currently cannot estimate the trajectory for the entire sequence. For higher resolution event cameras(346×260), we evaluate our methods on two sequences from MVSEC to demonstrate the applicability. Overall, the results of these complex motion sequences demonstrate the feasibility of our methods.
	
	Comparing the result of two proposed methods on the DAVIS dataset, it can be observed that the accuracy of GP-IMU is outperformed on most sequences due to its approximation of motion from IMU data. On sequences with drastic acceleration changes like sequence shapes\_6dof and boxes\_6dof, the results of two methods do not show consistent outcomes. The reason may be the aggressive motion seriously violates the WNOA prior in CT-IMU and also affects the 
	accuracy of IMU measurements to model motion which is important for GP-IMU to fuse asynchronous measurements. And for sequences from MVSEC where the IMU quality appears to be poor, the accuracy of two methods is not significantly different. Although the GP-IMU achieves satisfactory results, we observed that the GP-IMU method is more sensitive to IMU noise, as it relies on the IMU-driven GP to construct visual residuals. 
	\begin{figure}[!t]
		\vspace{8pt}
		\centering
		\includegraphics[width=3.2in]{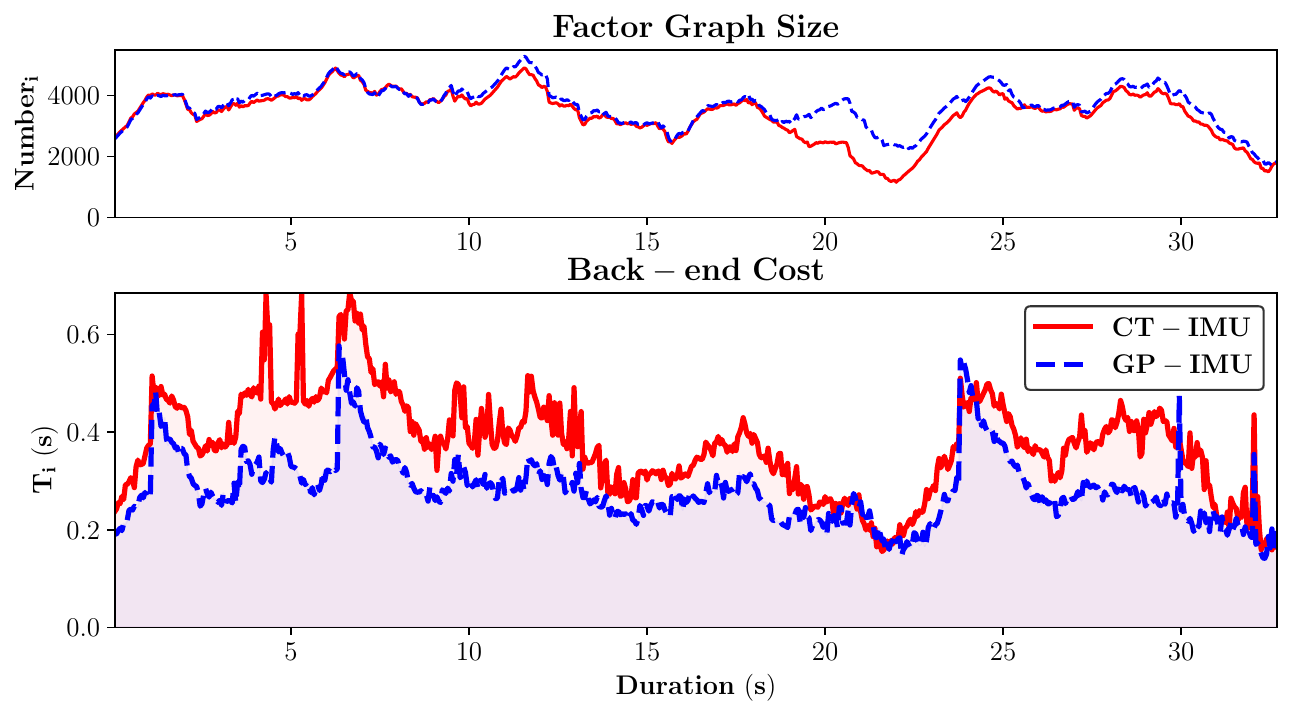}
		\caption{The factor graph's factor size and the each step ($\Delta t=0.05$s) back-end time cost of two methods on dynamic\_6dof sequence.}
		\label{dynamic_6dof_runtime}
	\end{figure}
	
	\begin{table*}[htb]
		\vspace{8pt}
		\caption{The Runtime of two methods on dynamic 6dof sequence.}
		\label{Runtime_table}
		\begin{center}
			\vspace{-2.0em}
			\begin{tabular}{ c | c | c | c | c | c}
				\hline\hline
				Method  &      front-end  & optimization & marginalization & IMU preintegration &  others \\
				\hline
				
				\textbf{CT-IMU}(s) &  1273.97 & 247.834 & \textbf{3.951} &  \textbf{0.177}  & 0.743  \\ 
				
				\textbf{GP-IMU}(s) & 1274.51 & \textbf{182.054} & 4.914 &  4.713 & \textbf{0.693}   \\
				\hline
			\end{tabular}
			\vspace{-2.0em}
		\end{center}
	\end{table*}
	\subsubsection{Time consumption}\label{consumption}
	We calculate the average time consumption of all parts of the system as shown in table \ref{Runtime_table}. The sequences we used have a duration approximately 35 seconds. The most time-consuming part is
	the EKLT tracker, which accounts for about $80\%$ of the total time. GP-IMU preintegration is slightly slower than standard IMU preintegration due to the calculation of intermediate latent states. For the whole graph optimization time, GP-IMU generally has a lower computational cost, which may be due to its fewer variables. Note that the dimension of state \eqref{state} in CT-IMU is $n\times 18 + m\times 3 $. In contrast, the GP-IMU's dimension is $n\times 15 + m\times 3 $. The time required for marginalization operations and other parts related to managing the factor graph does not  significantly differ between the two methods. However, with the marginalization, our system maintains a bounded computational cost, as shown in Fig. \ref{dynamic_6dof_runtime}. The number of factors in the two methods follows a very similar trend over time. The decrease of CT-IMU at 20 seconds may caused by the different success rates of triangulation between the two methods. Due to the fewer variables in GP-IMU and the smaller number of constraint variables for the landmark factor, GP-IMU performs faster than CT-IMU. 

\section{Conclusions}
In this work, we proposed two different GP-based ways to fuse asynchronous visual measurements and IMU measurements in event-based visual-initial odometry. The performance of two methods is quantitatively and qualitatively evaluated on several sequences. The results showed our methods achieve competitive performance in complex motion scenes compared with the state-of-the-art works. Since we kept all front asynchronous measurements in the optimization phase, the whole system cannot work real-time currently. However, we believe using a sparsification on measurement factors with our marginalization strategy will make our system to be real-time.  In the future, we will sparse our graph based on information theory and develop the front-end to make the whole system more robust.  

\bibliographystyle{IEEEtran}
\bibliography{IEEEabrv,paperbib}

\end{document}